\documentclass[letterpaper, 10 pt, conference]{ieeeconf}  % Comment this line out if you need a4paper

\IEEEoverridecommandlockouts                              % This command is only needed if 
\usepackage{fancyhdr}
\usepackage{graphics} % for pdf, bitmapped graphics files
\usepackage{graphicx}
\usepackage[dvipsnames]{xcolor}

\usepackage{booktabs} % Pour des tableaux plus esthétiques
\usepackage{array} % Pour les options de tableau supplémentaires

\usepackage{algorithmic}
\usepackage[linesnumbered,ruled,vlined]{algorithm2e}
\usepackage{amsfonts}
\usepackage{amsmath, xparse}
\usepackage{amssymb}
\usepackage{bm}
\usepackage{stmaryrd}

\usepackage{mathtools}

\makeatletter
\let\NAT@parse\undefined
\makeatother

\usepackage{url}
\usepackage[colorlinks=true,
            linkcolor=MidnightBlue,   % liens internes (sections, chapitres, figures…)
            citecolor=OliveGreen,     % citations biblio
            urlcolor=BrickRed,
            hyperfootnotes=false]
            {hyperref}

\usepackage[capitalize,nameinlink]{cleveref}
\crefname{figure}{Fig.}{Figs.}

\usepackage[frozencache, cachedir=./]{minted}
\DeclareFontFamily{U}{mathb}{\hyphenchar\font45}
\DeclareFontShape{U}{mathb}{m}{n}{
      <5> <6> <7> <8> <9> <10> gen * mathb
      <10.95> mathb10 <12> <14.4> <17.28> <20.74> <24.88> mathb12
      }{}
\DeclareSymbolFont{mathb}{U}{mathb}{m}{n}
\DeclareFontSubstitution{U}{mathb}{m}{n}

\let\dot\relax
\DeclareMathAccent{\dot}{0}{mathb}{"39}
\let\ddot\relax
\DeclareMathAccent{\ddot}{0}{mathb}{"3A}
\let\dddot\relax
\DeclareMathAccent{\dddot}{0}{mathb}{"3B}
\let\ddddot\relax
\DeclareMathAccent{\ddddot}{0}{mathb}{"3C}

\usepackage{acronym}
\makeatletter
\AtBeginDocument{%
  \@ifundefined{AC@hyperlink}{}{%
    \renewcommand*{\AC@hyperlink}[2]{\hyperlink{#1}{\textcolor{black}{#2}}}%
  }%
}
\makeatother

\title{\LARGE \bf
    PlaCo: a \acs{QP}-based robot planning and control framework
}

\author{Marc Duclusaud$^{1*}$, Grégoire Passault$^{1*}$, Vincent Padois$^{2}$, Olivier Ly$^{1}$% <-this % stops a space
\thanks{$^{1}$Univ. Bordeaux, CNRS, LaBRI, UMR 5800, 33400 Talence, France. Corresponding author: Marc Duclusaud, e-mail: \texttt{marc.duclusaud@u-bordeaux.fr}}
\thanks{$^{2}$Inria, Auctus, 33400 Talence, France.}
\thanks{$^{*}$Both authors contributed equally to this work.
\newline This study has received financial support from the French government in the framework of the France 2030 program, Initiative of Excellence (IdEx) University of Bordeaux / RRI ROBSYS.}
}

\begin{document}

\acrodef{CoM}{Center of Mass}
\acrodef{DoF}{Degree of Freedom}
\acrodef{IK}{Inverse Kinematics}
\acrodef{IMU}{Inertial Measurement Unit}
\acrodef{LIPM}{Linear Inverted Pendulum Model}
\acrodef{QDD}{Quasi-Direct Drive}
\acrodef{QP}{Quadratic Programming}
\acrodef{URDF}{Unified Robot Description Format}
\acrodef{WPG}{Walk Pattern Generator}
\acrodef{ZMP}{Zero Moment Point}

\maketitle
\thispagestyle{empty}
\pagestyle{empty}

%%%%%%%%%%%%%%%%%%%%%%%%%%%%%%%%%%%%%%%%%%%%%%%%%%%%%%%%%%%%%%%%%%%%%%%%%%%%%%%%
\begin{abstract}

This article introduces PlaCo, a software framework designed to simplify the formulation 
and solution of \ac{QP}-based planning and control problems for robotic systems. 
PlaCo provides a high-level interface that abstracts away the low-level mathematical 
formulation of \ac{QP} problems, allowing users to specify tasks and constraints 
in a modular and intuitive manner. The framework supports both Python bindings for rapid 
prototyping and a C++ implementation for real-time performance.

\end{abstract}

%%%%%%%%%%%%%%%%%%%%%%%%%%%%%%%%%%%%%%%%%%%%%%%%%%%%%%%%%%%%%%%%%%%%%%%%%%%%%%%%
\section{Introduction}

The control of robotic systems generally involves solving constrained optimization 
problems of considerable complexity. In addition, these problems must be solved 
under strict real-time requirements, since control loops typically operate at 
frequencies on the order of hundreds of hertz. A widely adopted approach 
to adress this challenge is to cast them as convex \acf{QP} problems, 
as they offer a certain number of desirable properties for robotic control.
First, they are convex, guaranteeing the existence of a unique 
solution that can be computed efficiently using one of the many 
available numerical solvers~\cite{qpsolvers}. Then, the optimization 
of a quadratic cost functions naturally promote smooth control commands. 
They also naturally support multiple objectives through weighted 
quadratic terms, allowing flexible prioritization. Finally, the implementation of 
constraints, such as joint bounds, velocity and torque limits, 
or contact stability constraints, can be included as linear 
inequalities and integrated into the problem.

Their standard formulation~\cite{nocedal2006quadratic, frank1956algorithm} is:
\begin{equation}
\begin{aligned}
& \underset{x}{\text{min}}
& & \dfrac{1}{2} x^T P x + a^T x \\
& \text{subject to}
& & Gx \leq h, \\
&&& Ax = b, 
\end{aligned}
\label{eq:qp}
\end{equation}
where $x$ denotes the decision variables to be optimized, 
$P$ and $a$ are 
the Hessian and gradient of the quadratic cost respectively, and $G$, $h$, 
$A$, $b$ define the linear inequality and equality constraints. 
The matrix $P$ is required to be positive semi-definite to ensure convexity.

The process of deriving the matrices for complex control problems 
can be tedious and error-prone. Moreover, this process
must be repeated whenever the control problem is modified or extended,
limiting rapid experimentation. Having a framework that abstracts away 
the low-level \ac{QP} formulation while allowing tasks and constraints to be 
specified in a simplified and modular manner is therefore highly desirable. 

\begin{figure}[t]
    \centering
    \includegraphics[width=0.48\textwidth]{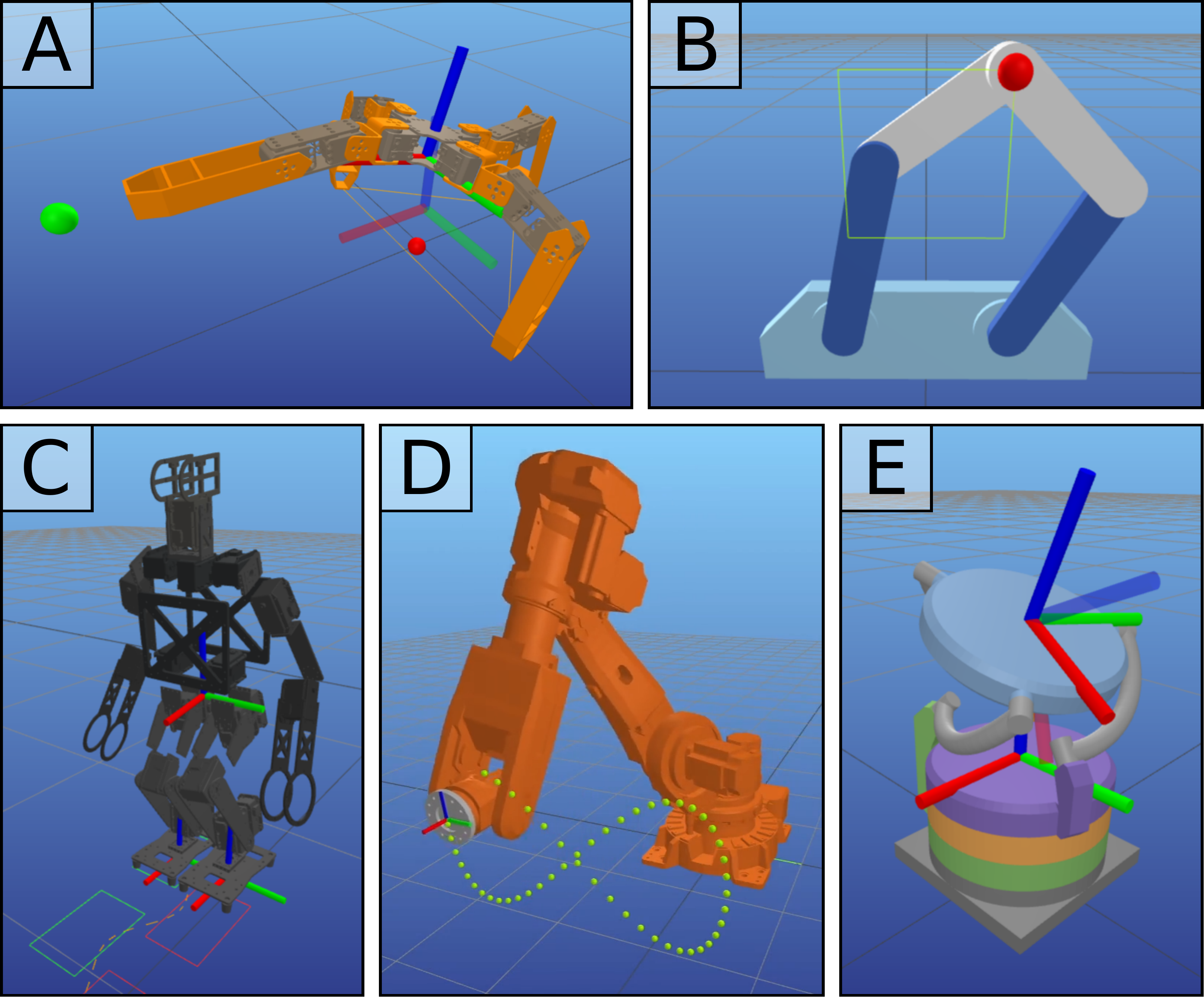}
    \vspace{-0.5em}
    \caption{Usage examples of the PlaCo \ac{IK} solver on several robots. All examples 
    are available in the example gallery\footref{fn:gallery}. 
    (\textbf{A}) A quadruped robot trying to reach the green target while strictly
    maintaining balance by enforcing the \ac{CoM} to stay inside the support polygon. 
    (\textbf{B}) A planar 2-\acsp{DoF} robot with kinematic loop following a square trajectory.
    (\textbf{C}) An humanoid robot walking following \acs{CoM}, footstep and pelvis orientation trajectories. 
    (\textbf{D}) A 6-\acsp{DoF} robotic arm following a trajectory.
    (\textbf{E}) An orbita actuator~\cite{crampette2020orbita} aiming at a target orientation while respecting joint limits.}
    \label{fig:catch-eye}
    \vspace{-1em}
\end{figure}

This work addresses this need by introducing \textbf{PlaCo} (Planning \& Control), a 
framework that simplify the formulation of planning and control 
problems as convex \ac{QP} problems for robotic systems. PlaCo provides 
a high-level specification layer that abstracts away low-level matrix 
assembly, Python bindings for interactive development and 
benchmarking, and a C++ implementation that ensures real-time performance. Rather 
than prescribing a fixed control architecture, PlaCo offers a set of composable 
building blocks for task specification, prioritization, and constraint handling. 
This flexibility allows to design a wide range of complex control tasks, such as 
the examples illustrated in \cref{fig:catch-eye}, which are taken from the
example gallery of the official PlaCo documentation\footnote{\label{fn:gallery}\href{https://placo.readthedocs.io/en/latest/kinematics/examples_gallery.html}
{placo.readthedocs.io/en/latest/kinematics/examples\_gallery.html}}.

The main contributions of PlaCo are:
\begin{itemize}
  \item \textbf{Abstraction of \ac{QP} formulation:} users can specify complex 
  constrained optimization problems for robot planning or control without the manual 
  construction of the underlying matrices.
  \item \textbf{Flexible \acf{IK} framework:} \ac{IK} problems can be constructed 
  from a wide range of task and constraint types -- including end-effector poses, 
  center of mass regulation, and relative poses -- with both hard and soft priorities.
  \item \textbf{Rapid prototyping (Python):} a concise API enables interactive 
  design, debugging, and evaluation of controllers and planners.
  \item \textbf{Real-time performance (C++):} the compiled core sustains the 
  control-loop frequencies required for whole-body task-space control.
\end{itemize}

The remainder of this article is organized as follows. 
First, \cref{sec:placo-architecture} presents the overall architecture of PlaCo. 
\cref{sec:problem-formulation} then details the problem abstraction provided by 
the framework. This abstraction is then applied to the design of a kinematics 
solver in \cref{sec:kinematics-solver}. Finally, \cref{sec:applications} illustrates
several applications to the control of robotic systems.

\section{PlaCo architecture}
\label{sec:placo-architecture}

\begin{figure}[!ht]
    \centering
    \includegraphics[width=0.48\textwidth]{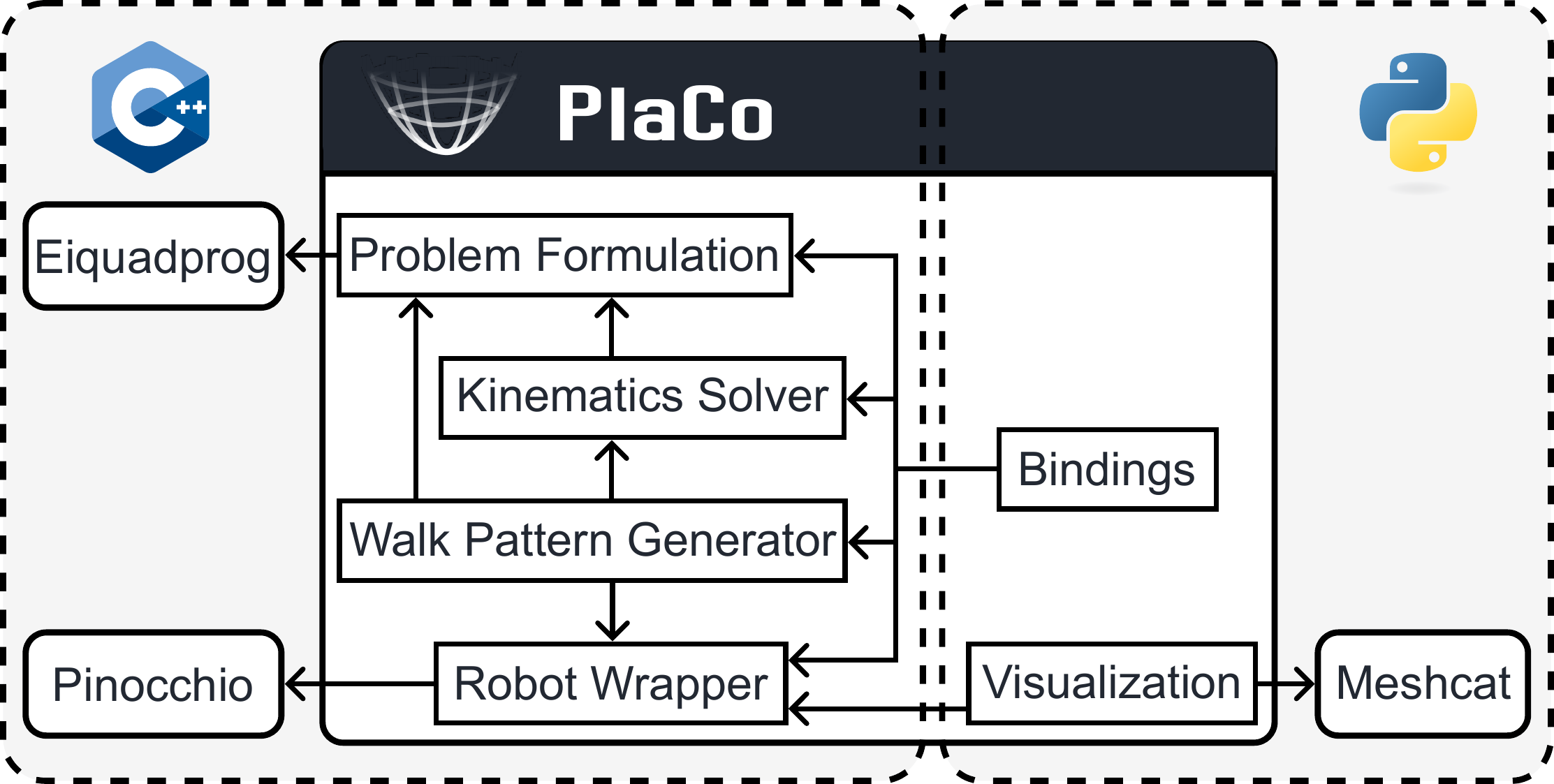}
    \vspace{-1em}
    \caption{Overview of the PlaCo architecture.}
    \label{fig:placo-architecture}
\end{figure}

The implementation of PlaCo is organized around two main components: 
(i) a high-level problem formulation module, and (ii) a dedicated kinematics solver. 
\cref{fig:placo-architecture} illustrates this structure, showing both the 
internal decomposition of PlaCo and its external dependencies, namely 
Pinocchio~\cite{carpentier2019pinocchio} for rigid-body dynamics, 
EiQuadProg~\cite{eiquadprog_soft, goldfarb1983numerically} for \ac{QP} solving, 
and Meshcat~\cite{meshcat_soft} for visualization. In addition, PlaCo includes a 
\acf{WPG} module specifically designed for humanoid locomotion. This module 
is still under development and will not be discussed further in this article. \\

\section{Problem abstraction}
\label{sec:problem-formulation}

The core principle of PlaCo is to provide a high-level interface for specifying
robot control problems, while internally reformulating them into the 
\ac{QP} formulation introduced in equation~\eqref{eq:qp} 
expected by efficient numerical solvers. This section illustrates how common robotics 
problems naturally reduce to this form.

First, \cref{sec:standard-qp} recalls the equivalence between least-squares 
objectives and the standard \ac{QP} formulation. \cref{sec:multi-obj} extends 
this formulation to the case of multiple objectives.  
\cref{sec:hard-soft} discusses how to incorporate hard and soft 
constraints into the \ac{QP} framework. \cref{sec:integrator} introduces integrated 
decision variables, which allow system dynamics to be embedded directly into the \ac{QP} problem. 
Finally, \cref{sec:qr-decomp} presents how QR factorization is used to 
reduce the dimensionality of the optimization problem. An usage example is 
provided in \hyperlink{appendix:usage-example}{Appendix A} to illustrate the 
problem specification process in PlaCo.

\subsection{From least-squares to standard \ac{QP} formulation}
\label{sec:standard-qp}

A least-squares minimization problem is formulated as 
\begin{equation}
    \min_{x} \; \|Mx - v\|^2,
    \label{eq:least-squares}
\end{equation}
where $x$ denotes the decision variables, $M$ a task-dependent matrix, 
and $v$ the target vector. This least-squares formulation directly minimizes 
the quadratic error between the linear function $Mx$ 
and the target $v$. 

Expanding this cost function reveals its equivalence 
to the standard \ac{QP} form presented in equation~(\ref{eq:qp}):
\begin{align}
    ||Mx - v||^2 &= (Mx - v)^T (Mx - v) \nonumber \\
                 &= (x^TM^T - v^T) (Mx - v) \\
                 &= x^TM^TM x - v^TMx - x^TM^Tv + v^T v \nonumber
\end{align}

Since $v^Tv$ is constant with respect to 
$x$, it can be ignored in the minimization.  
As $v^TMx = x^TM^Tv$, 
it follows that \eqref{eq:least-squares} is equivalent to:
\begin{equation}
    \underset{x}{\text{min}} \hspace{0.8em} \dfrac{1}{2} x^T M^T M x - v^TMx
\end{equation}

By defining $P = M^TM$ and 
$a = -M^Tv$, the standard 
\ac{QP} formulation introduced in equation~\eqref{eq:qp} is recovered. 
This is the form implemented by most numerical solvers commonly used 
in robotics applications, such as EiQuadProg \cite{goldfarb1983numerically}, qpOASES \cite{ferreau2014qpoases} 
or OSQP \cite{stellato2020osqp}.

\subsection{Multiple objectives problems}
\label{sec:multi-obj}

In most robotics applications, control problems rarely involve a single 
objective. Instead, multiple criteria must typically be optimized simultaneously, 
such as task tracking, balance maintenance, or collision avoidance.  
\ac{QP} provides a natural framework for this setting, as its 
additive quadratic cost structure readily accommodates multiple objectives.

Following the notations introduced earlier, and considering $k$ objectives 
weighted by positive coefficients $w_i$, the quadratic cost takes the form:
\begin{equation}
    \sum_{i=1}^k w_i \,\|M_i x - v_i\|^2, \qquad w_i>0
\end{equation}

The resulting optimization problem corresponds to the following \ac{QP} problem, where all objectives 
are aggregated into a single quadratic cost function:
\begin{equation}
    \underset{x}{\text{min}} \hspace{0.8em} \dfrac{1}{2} 
    x^T \left(\sum_{i=1}^{k} w_i M_i^\top M_i \right) x - 
     \left( \sum_{i=1}^{k} w_i v_i^\top M_i \right) x
\end{equation}

\subsection{Hard and soft constraints}  
\label{sec:hard-soft}

Within a \ac{QP} problem, constraints on the decision variables can be 
expressed as linear equalities or inequalities. The following defines 
two categories of constraints that differ in how strictly they are enforced.

Hard constraints are constraints that must be satisfied exactly. 
They are introduced as equality or inequality 
conditions in the solver through the matrices and vectors $A, b, G, h$ 
of the \ac{QP} formulation~\eqref{eq:qp}. 
For the problem to be feasible, all hard constraints must 
be mutually consistent; otherwise, the \ac{QP} problem becomes infeasible.

Soft constraints, on the other hand, may be violated at the cost of 
an additional penalty in the objective function. They are 
implemented differently depending on whether they are equalities or inequalities.  

A soft equality constraint is incorporated directly into 
the objective function as an additional quadratic term.  
For example, a constraint of the form
\begin{equation}
    Ax = b
\end{equation}
can be relaxed into the penalty
\begin{equation}
    w \|Ax - b\|^2    
\end{equation}
in the cost function, where $w>0$ is a weight determining 
the relative importance of the constraint. 

Soft inequality constraints require a more elaborate treatment, 
as a direct penalty on the violation would not 
distinguish between being inside or outside the feasible region.  
They are therefore relaxed by introducing auxiliary decision variables, 
known as \emph{slack variables}. 

Consider a constraint of the form
\begin{equation}
    Gx \leq h
    \label{eq:ineq}
\end{equation}

Defining this constraint as soft consists in extending the decision 
variables with a slack variable $s$, 
imposing the additional hard constraint $s \geq 0$, 
and adding to the cost function the term
\begin{equation}
    w \left\|
    G x - h + s
    \right\|^2
    \quad
    \Longleftrightarrow
    \quad
    w \left\|
    \begin{bmatrix}G & I\end{bmatrix} 
    \begin{bmatrix}
        x \\
        s
    \end{bmatrix} 
    - h
    \right\|^2
\end{equation}

With this formulation, as long as the original inequality $Gx \leq h$ 
is satisfied, the slack variable verify $s = h - Gx \ge 0$ 
and the penalty vanishes. When the inequality is violated, the hard 
constraint $s \geq 0$ forces $s=0$, 
and the penalty increases quadratically with the violation. 

It is worth noting that slack variables increase the 
dimensionality of the optimization problem, 
which may impact solver performance.  
Soft inequality constraints should therefore be used 
selectively and only when necessary.  

\subsection{Integrating decision variables}
\label{sec:integrator}

In many robotic systems, the temporal evolution of the state can be accurately 
described by a linear model. In such cases, the system dynamics can be directly 
embedded within the \ac{QP} formulation, enabling the optimization of the entire 
state trajectories over a finite horizon within a single problem.

Consider a system governed by linear continuous-time dynamics:
\begin{equation}
    \dot{y}(t) = D\,y(t) + E\,x(t),
\end{equation}
where $y(t) \in \mathbb{R}^m$ and $x(t) \in \mathbb{R}^p$ refers to the system 
state and control input at time $t$, and $D \in \mathbb{R}^{m \times m}$ and 
$E \in \mathbb{R}^{m \times p}$ are the continuous-time 
state and input matrices respectively.

The objective is to embed these dynamics into a \ac{QP} formulation with 
$x(t)$ as decision variables, so that constraints and objectives 
can be defined directly on the resulting state trajectory $y(t)$. 
As the \ac{QP} framework operates in discrete time, the control inputs are 
discretized over a prediction horizon of $N$ time steps with sampling period $\Delta t$:
\begin{equation}
    x =
    \begin{bmatrix}
    x_0 & x_1 & \ldots & x_{N-1}
    \end{bmatrix}^T
\end{equation}

The corresponding discrete-time dynamics are
\begin{equation}
    y_{k+1} = D_{d}\,y_k + E_{d}\,x_k,
    \label{eq:discrete-time}
\end{equation}
where $y_k \in \mathbb{R}^m$ and $x_k \in \mathbb{R}^p$ denote the 
system state and control input at time step $k$, and $D_{d} \in \mathbb{R}^{m \times m} $ 
and $E_{d} \in \mathbb{R}^{m \times p}$ are the
discrete-time state and input matrices respectively.

The matrices $D_{d}$ and $E_{d}$ are derived 
from the continuous-time dynamics using the exact discretization via block-matrix 
exponentials~\cite{van2003computing}:
\begin{equation}
    \exp\!\left(
    \begin{bmatrix}
        D & E \\
        0 & 0
    \end{bmatrix}
    \Delta t
    \right)
    =
    \begin{bmatrix}
        D_{d} & E_{d} \\
        0 & I
    \end{bmatrix}
    \label{eq:discrete-matrices}
\end{equation}

By recursively applying the discrete-time dynamics 
\eqref{eq:discrete-time} from the initial state $y_0$, 
the state at step $k$ can be expressed as:
\begin{equation}
\begin{aligned}
    y_1 &= D_{d}\,y_0 + E_{d}\,x_0 \\
    y_2 &= D_{d}^2\,y_0 + D_{d}\,E_{d}\,x_0 + E_{d}\,x_1 \\
    y_3 &= D_{d}^3\,y_0 + D_{d}^2\,E_{d}\,x_0 + D_{d}\,E_{d}\,x_1 + E_{d}\,x_2 \\
    &\vdots \\
    y_k &= D_{d}^k\,y_0 +
    \sum_{i=0}^{k-1} D_{d}^{k-1-i}\,E_{d}\,x_i
\end{aligned}
\end{equation}

% Equivalently, in compact matrix form:
% \begin{equation}
% \forall k \in \llbracket 1, N \rrbracket, \quad
% y_k = D_{d}^k y_0 +
% \begin{bmatrix}
%     D_{d}^{k-1} E_{d} &
%     D_{d}^{k-2} E_{d} &
%     \ldots &
%     D_{d} E_{d} &
%     E_{d} &
%     \smash{\underbrace{0 \ \ldots \ 0}_{N-k \text{ terms}}}
% \end{bmatrix}
% x
% \end{equation}

Thus, given the initial state $y_0$ and the discrete-time
state and input matrices $D_{d}$ and $E_{d}$,
one can formulate linear constraints on $y_k$ at any time
step $k$ as a function of the decision variables $x$. 
The matrices involving powers of $D_{d}$ are precomputed 
and reused to improve computational efficiency when 
solving the \ac{QP} problem repeatedly.

Finally, to express constraints on the continuous-time state $y(t)$ 
at arbitrary instants $t \in [0, N \Delta t]$, the continuous-time dynamics 
can be used to propagate the state from the last discrete step preceding $t$. 
Specifically, for $t \in [k \Delta t, (k+1) \Delta t]$, 
the corresponding state and input matrices for a duration 
$\tau = t - k \Delta t$ are defined, using the same formulation as in 
\eqref{eq:discrete-matrices}:
\begin{equation}
    \exp\!\left(
    \begin{bmatrix}
        D & E \\
        0 & 0
    \end{bmatrix}
    \tau
    \right)
    =
    \begin{bmatrix}
        D_{d}^\prime & E_{d}^\prime \\
        0 & I
    \end{bmatrix}
    \label{eq:discrete-matrices-prim}
\end{equation}

The continuous-time state at time $t$ is then given by:
\begin{equation}
    y(t) = D_{d}^\prime y_k +
    E_{d}^\prime x_k
\end{equation}

This allows constraints on the state of the system to be defined at arbitrary time points within the
prediction horizon, while remaining computationally efficient, 
as only the matrix exponential in \eqref{eq:discrete-matrices-prim} 
must be computed online. Examples to illustrate this approach are provided 
in \hyperlink{appendix:integrated-example}{Appendix B}.

\subsection{Reducing problem dimensionality with QR factorization}
\label{sec:qr-decomp}

The dimensionality and the number of constraints of a \ac{QP} problem 
have a direct impact on the computational cost of its solution.
Therefore, reducing the number of decision variables and limiting 
the number of constraints is essential to achieve real-time performance. 
Bemporad \textit{et al.}~\cite{bemporad2021variable} proposed a method 
that employs QR factorization to simultaneously eliminate all hard equality 
constraints from the \ac{QP} formulation and reduce the number of decision 
variables by the number of removed constraints.

Consider a constrained least-squares problem of the form:
\begin{equation}
    \begin{aligned}
    & \underset{x}{\text{min}} & & 
    ||Mx - v||^2 \\
    & \text{subject to} & & Ax = b, \\
    & & & Gx \leq h,
    \end{aligned}
    \label{eq:constrained-ls}
\end{equation}
where $x \in \mathbb{R}^p$ are the decision variables, 
and $A \in \mathbb{R}^{r \times p}$ is the equality constraint 
matrix. The hard equality constraints being generally independent,
the matrix $A$ is assumed to have full row rank $r$.

% As detailed in \cref{sec:standard-qp}, 
% this problem can be reformulated as a 
% \ac{QP} problem of the form introduced in equation~\eqref{eq:qp} by defining 
% $P = M^TM$ and 
% $a = -M^Tv$.

The full QR factorization of $A^T$ can be written as:
\begin{equation}
    A^T = QR, 
    \quad Q = \begin{bmatrix} Q_1 & Q_2 \end{bmatrix},
    \quad R = 
    \begin{bmatrix}
    R_1 \\
    0
    \end{bmatrix},
\end{equation}
where $Q \in \mathbb{R}^{p \times p}$ is orthogonal, 
$Q_1 \in \mathbb{R}^{p \times r}$, $Q_2 \in \mathbb{R}^{p \times (p-r)}$,
$R \in \mathbb{R}^{p \times r}$, and $R_1 \in \mathbb{R}^{r \times r}$ 
is upper-triangular. Note that since $A$ has rank $r$, 
$R_1$ is nonsingular.

By introducing the following change of variables:
\begin{equation}
    \bar{x} = Q_2^T x,
\end{equation}
the constrained problem in~\eqref{eq:constrained-ls} is equivalent to the reduced problem:
\begin{equation}
    \begin{aligned}
    & \underset{\bar{x}}{\text{min}} & & 
    \big\| MQ_2\bar{x} - 
    \big(v - MQ_1(R_1^T)^{-1}b\big) \big\|^2 \\
    & \text{subject to} & & 
    GQ_2 \bar{x} \leq 
    h - GQ_1(R_1^T)^{-1}b.
    \end{aligned}
\end{equation}

This reduced problem has $p - r$ decision variables and no equality constraints.
A formal proof of this equivalence can be found in Section~II-A of~\cite{bemporad2021variable}.

Although the change of variables itself introduces a computational overhead, 
tests conducted with PlaCo demonstrate that the overall performance 
is consistently improved by applying this reduction. The benefit becomes 
more significant as the number of the equality constraints increases, 
since this leads to a larger reduction in problem size.

\section{Kinematics solver}
\label{sec:kinematics-solver}

The kinematics solver of PlaCo builds upon the problem formulation 
introduced in \cref{sec:problem-formulation}.
It provides a whole-body framework for inverse kinematics, where 
tasks and constraints can be specified directly in task space.
Both soft and hard priorities are supported, enabling flexible 
combinations of objectives while ensuring the respect of essential 
feasibility conditions, such as joint position and velocity limits.

The remainder of this section is organized as follows.
\cref{sec:ik-formulation} introduces the task-space 
\ac{IK} formulation that underlies the solver.
\cref{sec:tasks} reviews the main categories of 
tasks available in PlaCo, while \cref{sec:constraints} 
presents the principal classes of constraints.

\subsection{Task-space \ac{IK} problem formulation}
\label{sec:ik-formulation}

Consider a serial articulated robot with $n$ actuated joints.
The robot configuration is fully described by the vector
$q \in \mathbb{R}^n \times \mathbb{SE}(3)$, which consists of the position and
orientation of the floating base $q_u \in \mathbb{SE}(3)$ and the joint positions
$q_a \in \mathbb{R}^n$:
\begin{equation}
    q =
    \begin{bmatrix}
        q_u \\
        q_a
    \end{bmatrix}
\end{equation}

The current configuration of the robot is denoted by $q_0$.
The task-space \ac{IK} problem is defined as the determination of a configuration
increment $\Delta q$ such that
\begin{equation}
    q = q_0 + \Delta q
\end{equation}
satisfies a set of tasks, such as positioning a rigid body at 
a desired pose, and simultaneously fulfills a set of constraints, 
for instance ensuring that the center of mass remains within the support polygon. 
As configuration spaces involving rotations are not Euclidean, the 
floating base configuration $q_u$
update is generally not a strict vector addition. 
Nevertheless, the same principle applies, and the simplified
additive notation is used here for clarity.

This problem naturally translates into a constrained optimization setting,
where the configuration increment $\Delta q$ serves as the
decision variable.

For each task $i$, an error function $e_i$ is defined as the deviation from
the desired task value. A task is perfectly satisfied when
$e_i(q) = 0$. For sufficiently small
$\Delta q$, a first-order Taylor expansion around the current
configuration $q_0$ yields
\begin{equation}
    e_i(q_0 + \Delta q)
    \approx e_i(q_0) +
    J_i(q_0)\, \Delta q,
\end{equation}
where $J_i(q_0)$ is the Jacobian of the task error $e_i$ 
evaluated at $q_0$. \\

Two categories of tasks are typically distinguished:
\begin{itemize}
    \item \textbf{Hard-priority tasks}, which must be satisfied exactly and
    are expressed as hard equality constraints:
    \begin{equation}
        e_i^h(q_0) + J_i^h(q_0)\, \Delta q = 0,
    \end{equation}

    \item \textbf{Soft-priority tasks}, which are weighted by coefficients
    $w_i > 0$ and penalized quadratically in the objective:
    \begin{equation}
        w_i \, \|e_i^s(q_0) + J_i^s(q_0)\, \Delta q\|^2
    \end{equation}
\end{itemize}

In addition to tasks, which are equality-driven, the \ac{IK} problem also involves 
constraints that restrict the feasible configuration space.  
Formally, a constraint $j$ is represented by a function 
$g_j$, which must remain non-positive to be satisfied.
For sufficiently small $\Delta q$, this condition is approximated 
by its first-order expansion at $q_0$:
\begin{equation}
    g_j(q_0) + G_j(q_0)\, \Delta q \leq 0,
\end{equation}

where $G_j(q_0)$ is the Jacobian of the constraint
function $g_j$ evaluated at $q_0$. \\

As with tasks, two categories of constraints can be defined:
\begin{itemize}
    \item \textbf{Hard-priority constraints}, which must be satisfied exactly and
    are expressed as hard inequality constraints:
    \begin{equation}
        g_j^h(q_0) + G_j^h(q_0)\, \Delta q \leq 0,
    \end{equation}

    \item \textbf{Soft-priority constraints}, which may be violated at a penalized cost.
    As presented in \cref{sec:hard-soft}, this is handled by introducing 
    slack variables $s_j \geq 0$ and adding to the objective penalties
    weighted by coefficients $\rho_j > 0$:
    \begin{equation}
        \rho_j \,\| g_j^s(q_0) + G_j^s(q_0)\, 
        \Delta q + s_j \|^2
    \end{equation}
\end{itemize}

Finally, to improve numerical conditioning, a small regularization term is
added to the objective:
\begin{equation}
    \epsilon \,\|\Delta q\|^2, \quad \forall i, w_i \gg \epsilon > 0
\end{equation}

This regularization term is equivalent to a soft task that minimizes the norm of the configuration
increment, weighted by a small coefficient $\epsilon$.

Considering a set of $m^h$ hard tasks, $m^s$ soft tasks, $l^h$ hard constraints,
and $l^s$ soft constraints, the complete \ac{IK} problem can therefore be expressed as:
\begin{align}
    & \underset{\Delta q}{\text{min}}
    & & \sum_{i=1}^{m^s} w_i \,\| e_i^s(q_0) + J_i^s(q_0) \Delta q\|^2 \nonumber \\
    &&& + \sum_{j=1}^{l^s} \rho_j \,\| g_j^s(q_0) + G_j^s(q_0)\, \Delta q + s_j \|^2 \nonumber \\[2ex]
    &&& + \epsilon \,\|\Delta q\|^2 \label{eq:ik_qp}
\end{align}
\begin{equation*}
    \begin{aligned}
    & \text{subject to}
    & & \forall i \in \llbracket 1, m^h \rrbracket, &\text{} &e_i^h(q_0) + J_i^h(q_0) \Delta q = 0, \\
    &&& \forall j \in \llbracket 1, l^h \rrbracket, &\text{} &g_j^h(q_0) + G_j^h(q_0)\, \Delta q \leq 0, \\
    &&& \forall j \in \llbracket 1, l^s \rrbracket, &\text{} &s_j \geq 0
    \end{aligned}
\end{equation*}

This \ac{QP}-based formulation can be interpreted as a single step of a constrained Gauss--Newton method; 
when iterated, it effectively implements a sequential \ac{QP} procedure as defined in~\cite{nocedal2006quadratic}.

\subsection{Main categories of tasks}
\label{sec:tasks}

As introduced in the previous section, each task $i$ is expressed as an error function 
that measures the deviation between the current configuration and the desired 
task objective. Linearizing this error around $q_0$ yields: 
\begin{equation}
    e_i(q) \approx e_i(q_0) + J_i(q_0)\,\Delta q
\end{equation}

From the solver’s perspective, all tasks therefore reduce to a linear function 
of the configuration increment to minimize. Internally, task $i$ is represented by the pair 
$(J_i(q_0), e_i(q_0))$, which can either be 
enforced as a hard equality constraint in the \ac{QP} problem or penalized as a soft objective term.

Tasks can be freely combined within the solver, allowing the 
definition of control objectives such as joint positions, rigid body placement, 
or center of mass control. The error function and Jacobian associated with each task 
are derived from the underlying rigid body kinematics. In practice, these 
quantities are computed using efficient rigid body dynamics algorithms 
implemented in the Pinocchio library~\cite{carpentier2019pinocchio}.

The following subsections introduce the main task categories implemented in PlaCo. 
While not exhaustive, these represent the task formulations most frequently used in robotics.
For each case, the corresponding error function and Jacobian are defined, and the way 
they are implemented is detailed. \\~\vspace{-0.3em}

\subsubsection{Position and orientation tasks} ~\\~\vspace{-0.9em}
\label{sec:frame-task}

A \textbf{position task} constrains the translation of a rigid body or a frame attached 
to the robot with respect to the world frame, while an \textbf{orientation task} constrains 
its rotation. Both types of tasks can be used independently or combined 
into a \textbf{frame task} to form a full 6D constraint. These tasks are 
essential for defining end-effector objectives, such as controlling the 
feet during walking or positioning a hand at a desired location in a manipulation task. 

Let ${}^wp \in \mathbb{R}^3$ denote the 
translation of the frame and ${}^wR \in SO(3)$ its 
orientation in the world. Translations and rotations 
depend on the robot configuration $q_0$, but this 
dependence is omitted for readability throughout the paper.
Given a target translation ${}^wp_{\text{target}}$ 
and a target orientation ${}^wR_{\text{target}}$, the position task is defined by
\begin{equation}
    e_{\text{pos}}(q_0) = {}^wp_{\text{target}} - {}^wp,
    \text{ }
    J_{\text{pos}}(q_0) = {}^wJ^t_{\text{frame}}(q_0),
\end{equation}
while the orientation task is given by
\begin{equation}
    e_{\text{ori}}(q_0) = \mathrm{Log}\!\left(
        \,{}^wR_{\text{target}}^T \,{}^wR
    \right),
    \text{ }
    J_{\text{ori}}(q_0) = {}^wJ^r_{\text{frame}}(q_0)
\end{equation}

While combining position and orientation tasks into a frame task, 
the corresponding error components are kept decoupled rather than 
stacked into a 6D screw vector. This design choice is motivated, first, 
by improved interpretability, as the errors retain clear physical units -- 
meters for position and radians for orientation. Second, it guarantees 
that the motion between two frames with different orientations corresponds 
to a straight-line trajectory in Cartesian space, rather than the helical 
path that would result from a 6D screw interpolation. This property is 
generally desirable in robotics applications, as discussed in section~9.2.1 
of~\cite{lynch2017modern}.

To improve task versatility, PlaCo supports component-wise masking of 
error across all task types. 
% For frame tasks, this feature allows selectively constraining specific axes of translation or rotation. 
For instance, a position task can control only the $x$ and $y$ coordinates 
of a frame while leaving the $z$ axis unconstrained. Similarly, an orientation 
task can regulate only the yaw angle, with roll and pitch remaining free. 

\subsubsection{Relative position and orientation tasks}~\\~\vspace{-1em}
\label{sec:relative-tasks}

In addition to absolute tasks expressed with respect to the world frame, 
\textbf{relative position tasks} and \textbf{relative orientation tasks} 
can also be defined, as well as their combination into a \textbf{relative frame task}. 
These tasks constrain the pose of a rigid body or frame attached 
to the robot with respect to another frame of the robot. They are particularly useful for 
specifying relative positioning objectives between two end-effectors or for enforcing kinematic 
loop-closure constraints, as presented in \cref{sec:closed-loop}.

Let $^wp_a \in \mathbb{R}^3$ and 
$^wR_a \in SO(3)$ denote the translation and orientation of a 
frame $a$ expressed in the world frame, and similarly 
$^wp_b$, $^wR_b$ for another frame $b$.  
The relative translation of $b$ with respect to $a$ is then
\begin{equation}
    \,^ap_b = \,^wR_a^T
    \left(\,^wp_b - \,^wp_a \right),
\end{equation}
and the relative orientation is given by
\begin{equation}
    \,^aR_b =
    \,^wR_a^T \,^wR_b
\end{equation}

Given a target relative translation $\,^ap_{b,\text{target}}$ and a target 
relative orientation $\,^aR_{b,\text{target}}$, the corresponding errors are defined as
\begin{equation}
    e_{\text{pos}}^{b|a}(q_0) = \,^ap_{b,\text{target}} - \,^ap_b,
    \text{ }
    J_{\text{pos}}^{b|a}(q_0) = {}^aJ_b^t(q_0),
\end{equation}
\begin{equation}
    e_{\text{ori}}^{b|a}(q_0) = \mathrm{Log}\!\left(
        \,^aR_{b,\text{target}}^T \,^aR_b
    \right),
    \text{ }
    J_{\text{ori}}^{b|a}(q_0) = {}^aJ_b^r(q_0)
\end{equation}
\vspace{0em}
\subsubsection{\ac{CoM} task} ~\\~\vspace{-1em}
\label{sec:com-task}

A \textbf{\ac{CoM} task} regulates the position of the \ac{CoM} of the robot to a desired target.
This task is fundamental for balance control and plays 
a central role in locomotion scenarios, where stability 
must be preserved while the robot is in motion. 

Let $c \in \mathbb{R}^3$ denote the \ac{CoM} position. Given a target \ac{CoM} position 
$c_{\text{target}}$, the task error and its associated Jacobian are defined as
\begin{equation}
    e_{\text{com}}(q_0) = c_{\text{target}} - c,
    \quad
    J_{\text{com}}(q_0),
\end{equation}
where $J_{\text{com}}(q_0)$ is the \ac{CoM} Jacobian,
computed using the Pinocchio library~\cite{carpentier2019pinocchio}. \\~\vspace{-0.5em}

\subsubsection{Joint task} ~\\~\vspace{-1em}
\label{sec:joint-task}

A \textbf{joint task} aims to regulate the position of one or more actuated joints 
to specified target values. This is useful for maintaining a desired posture 
or for controlling redundant degrees of freedom. It is the most elementary 
task type, as its error function is simply the deviation between the current 
and desired joint positions.

Let $q_{\text{target}} \in \mathbb{R}^k$ denote the desired positions 
for a subset of $k$ joints, and let $S \in \mathbb{R}^{k \times (n+6)}$ be the selection 
matrix extracting from the configuration vector the components corresponding to the controlled joints. 
The task error and its associated Jacobian then take the form
\begin{equation}
    e_{\text{joint}}(q_0) = q_{\text{target}} - S q_0,
    \qquad
    J_{\text{joint}} = S
\end{equation}

\subsubsection{Gear task} ~\\~\vspace{-1em}
\label{sec:gear-task}

A \textbf{gear task} enforces a linear coupling between a target joint
and one or several source joints, such that their velocities are related
by constant ratios. This formulation captures the effect of mechanical transmissions such as
gears, timing belts, or cable-driven systems. It also naturally extends to 
more complex mechanisms, for instance differentials, as presented in \cref{sec:differential}, 
where the motion of one joint results from a linear combination of multiple sources. More generally,
it applies to any situation where one degree of freedom is required to reproduce
the motion of others, possibly scaled by coefficients. 

Consider a set of $k$ joints of the robot to be controlled through a gear
task. Let $S \in \mathbb{R}^{k \times (n+6)}$ denote 
the corresponding selection matrix, which extracts from the configuration vector 
the positions of these controlled joints. 

In addition, a weight matrix $W \in \mathbb{R}^{k \times (n+6)}$
is introduced, whose non-zero entries specify the linear combination 
of source joints that each controlled joint is expected to follow.
For each controlled joint $i \in \llbracket 1, k \rrbracket$,
let $\mathcal{S}_i \subseteq \llbracket 1, n+6 \rrbracket$ denote the set of source joint indices
that influence joint $i$, and let $w_{ij} \in \mathbb{R}$ be the 
ratio associated with source joint $j \in \mathcal{S}_i$.
The entries of $W$ are then given by
\begin{equation}
W_{ij} =
\begin{cases}
w_{ij} & \text{if } j \in \mathcal{S}_i \\
0 & \text{otherwise}
\end{cases}
\end{equation}

The gear task error then enforces consistency between the actual motion
of the controlled joints and their prescribed linear combination of sources:
\begin{equation}
e_{\text{gear}}(q_0) = S q_0 - W q_0,
\qquad
J_{\text{gear}} = S - W
\end{equation}

\subsection{Main categories of constraints}
\label{sec:constraints}

In the task-space \ac{IK} formulation, constraints complement tasks by restricting
the feasible configuration space of the robot. While tasks enforce equalities
that drive the robot toward desired objectives, constraints impose inequalities
that guarantee safety, feasibility, or physical consistency. Linearized around
$q_0$, each constraint $j$ is represented by the pair
$(G_j(q_0), g_j(q_0))$, leading to the
inequality
\begin{equation}
g_j(q_0) + G_j(q_0) \Delta q \leq 0
\end{equation}

As for tasks, constraints can be enforced with different levels of priority.
Hard constraints must be satisfied exactly and appear directly in the feasible
set of the \ac{QP} problem, whereas soft constraints are relaxed through slack variables
and penalized in the objective.

PlaCo implements a number of constraint formulations that address the main
requirements of whole-body control in robotics. These include, for instance,
constraints ensuring contact stability or avoiding self-collisions. 
While not exhaustive, the following subsections present the most 
representative constraint categories provided by the solver.

\subsubsection{Range constraints} ~\\~\vspace{-1em}
\label{sec:position-limits}

In every robotic system, physical limitations restrict the admissible 
range of motion of the joints. These bounds stem from mechanical design 
choices and must be incorporated into the control problem through 
\textbf{range constraints} to ensure that 
the computed motions are physically realizable. They also serve as a 
safety requirement, since commanding configurations beyond the allowed 
range may damage the robot.

Let $q_{\min} \in \mathbb{R}^{n+6}$ and
$q_{\max} \in \mathbb{R}^{n+6}$ denote the minimum and maximum
joint positions, respectively. Feasible configurations must satisfy
\begin{equation}
    q_{\min} \leq q \leq q_{\max}
\end{equation}

Linearizing around the current configuration $q_0$ yields 
two inequality constraints on the configuration increment:
\begin{equation}
    \begin{aligned}
    q_{\min} - q_0 - \Delta q \leq 0, \\
    q_0 - q_{\max} + \Delta q \leq 0
    \end{aligned}
\end{equation}

In canonical form, these constraints are expressed as
\begin{equation}
  \begin{aligned}
    g_{\text{range,min}}(q_0) &= q_{\text{min}} - q_0,
    &\quad G_{\text{range,min}}(q_0) &= -I, \\
    g_{\text{range,max}}(q_0) &= q_0 - q_{\text{max}},
    &\quad G_{\text{range,max}}(q_0) &= I
  \end{aligned}
\end{equation}

These bounds are typically enforced as hard constraints to guarantee 
that the robot remains within its physical limits. \\~\vspace{-0.5em}

\subsubsection{Velocity constraints} ~\\~\vspace{-1em}
\label{sec:velocity-limits}

Actuators in robotic systems have finite capabilities, which limit the speed
at which joints can move. To ensure that the computed motions are feasible
and respect these physical limitations, it is essential to include 
\textbf{velocity constraints} in the control problem. 

Let $\dot{q}_{\max} \in \mathbb{R}^{n+6}$ denote the maximum
joint velocities. Assuming a control time step $\Delta t$, the configuration
increment $\Delta q$ must satisfy
\begin{equation}
    -\dot{q}_{\max} \Delta t \leq \Delta q \leq \dot{q}_{\max} \Delta t
\end{equation}

Equivalently, this can be written as the pair of inequalities
\begin{equation}
    \begin{aligned}
    -\dot{q}_{\max} \Delta t - \Delta q \leq 0, \\
    -\dot{q}_{\max} \Delta t + \Delta q \leq 0
    \end{aligned}
\end{equation}

Expressed in canonical form, they read
\begin{equation}
    \begin{aligned}
    g_{\text{vel,min}}(q_0) &= -\dot{q}_{\max} \Delta t,
    &\quad G_{\text{vel,min}}(q_0) &= -I, \\
    g_{\text{vel,max}}(q_0) &= -\dot{q}_{\max} \Delta t,
    &\quad G_{\text{vel,max}}(q_0) &= I
    \end{aligned}
\end{equation}

Such bounds are usually enforced as hard constraints to ensure that 
the robot operates within its dynamic capabilities. In scenarios where 
the \ac{IK} solver is only used to generate reference configurations, they may 
be relaxed to accelerate convergence.

It should be noted that joint velocity and torque limits are coupled: 
faster motions are feasible under lighter loads, while high torques 
restrict achievable velocities. The present formulation does not capture 
this coupling, but it prevents unrealistic joint speeds that would be 
infeasible regardless of applied torques. \\~\vspace{-0.5em}

\subsubsection{Polygonal constraints} ~\\~\vspace{-1em}
\label{sec:polygonal-constraints}

Common stability criteria in legged locomotion often require ensuring that a 
two-dimensional quantity of interest, such as the horizontal projection 
of the \ac{CoM} of the robot, remains within a convex 
polygonal region on the ground plane. This type of constraint is referred 
to as a \textbf{polygonal constraint}. By not directly controlling the 
quantity but instead restricting it to a feasible region, they provide 
flexibility while ensuring that basic stability requirements are satisfied.

Formally, let $y \in \mathbb{R}^2$ denote the horizontal 
projection of the controlled quantity and $J_y(q_0) \in \mathbb{R}^{2 \times (n+6)}$ 
its Jacobian with respect to the configuration $q_0$. 

The polygonal region in which $y$ must remain
is defined by its vertices $V = \{ V_1, \dots, V_N \} \subset \mathbb{R}^2$,  
ordered clockwise. Indices are taken modulo $N$, so that $V_{N+1} = V_1$.  
For each edge $(V_i, V_{i+1})$, a unit inward normal vector $n_i \in \mathbb{R}^2$ 
is defined as:
\begin{equation}
    n_i = 
    \frac{1}{\|V_{i+1} - V_i\|} 
    \begin{bmatrix}
        0 & 1 \\
        -1 & 0
    \end{bmatrix}
    (V_{i+1} - V_i)
\end{equation}

Enforcing $y$ to remain inside the polygon with 
a safety margin $d_{\text{min}} \ge 0$ is equivalent to requiring
\begin{equation}
    \forall i \in \llbracket 1, N \rrbracket, 
    \quad 
    n_i^T \left( y - V_i \right) \geq d_{\text{min}}
\end{equation}

For each edge $i$, this condition can be linearized and rearranged into the canonical
inequality form:
\begin{equation}
    \underbrace{d_{\text{min}} - n_i^T \left( y - V_i \right)}_{g_i(q_0)} 
    + \underbrace{(-n_i^T) J_y(q_0)}_{G_i(q_0)} 
    \Delta q \leq 0 
\end{equation}

Stacking the inequalities for all polygon edges gives the global \ac{QP} constraint
\begin{equation}
    g_{\text{poly}}(q_0) 
    + G_{\text{poly}}(q_0) \, \Delta q \leq 0,
\end{equation}
where $g_{\text{poly}}(q_0) \in \mathbb{R}^N$ and 
$G_{\text{poly}}(q_0) \in \mathbb{R}^{N \times (n+6)}$ 
are obtained by stacking the contributions of all polygon edges. \\~\vspace{-0.5em}

% Several important stability constraints can be expressed with this formulation. 
% Choosing $y$ as the \ac{CoM} projection yields the \textbf{\ac{CoM} polygonal constraint},  
% illustrated in \cref{sec:quadruped-example} and particularly relevant for static balancing.  
% Alternatively, selecting $y$ as the \ac{ZMP} leads to the \textbf{\ac{ZMP} polygonal constraint}  
% employed in the walking pattern generator described in \cref{chap:walk}. 

% Alternatively, a similar formulation is employed in the \ac{WPG} described in \cref{chap:walk} 
% to constrain the \ac{ZMP} within the support polygon.

\subsubsection{Self-collision avoidance constraint} ~\\~\vspace{-1em}
\label{sec:self-collision}

A \textbf{self-collision avoidance constraint} prevents collisions between
different parts of the robot. This is crucial for ensuring safe and feasible
motions, especially in compact configurations or when operating in cluttered
environments. 

The constraint enforces that the distance between pairs of 
rigid bodies remains above a prescribed safety margin $d_{\min} > 0$.  
Let $(a,b)$ denote a collision pair of robot bodies.  
For a current configuration $q_0$, the coal library~\cite{coalweb} 
allows to determine the coordinates ${}^wp_A, {}^wp_B \in \mathbb{R}^3$ 
of the closest points $A$ and $B$ of the two bodies expressed in the world frame, 
together with their signed minimal distance $d_{AB}(q_0) \in \mathbb{R}$.  
By convention, this distance is positive when the bodies are 
separated and negative when they interpenetrate.

To linearize the constraint, a unit vector ${}^wn_{AB}$ 
is defined so that it consistently points in the direction that increases 
the separation of body $b$ with respect to body $a$:
\begin{equation}
    {}^wn_{AB} =
    \begin{cases}
        \dfrac{{}^wp_B - {}^wp_A}{\|{}^wp_B - {}^wp_A\|} & \text{if } d_{AB}(q_0) \geq 0 \\[3ex]
        \dfrac{{}^wp_A - {}^wp_B}{\|{}^wp_A - {}^wp_B\|} & \text{if } d_{AB}(q_0) < 0
    \end{cases}
\end{equation}

The linearized distance constraint then reads
\begin{equation}
    d_{AB}(q_0) 
    + {}^wn_{AB}^T 
    \left( {}^wJ^t_B(q_0) - {}^wJ^t_A(q_0)\right) 
    \Delta q 
    \geq d_{\min},
\end{equation}
where ${}^wJ^t_A(q_0)$ and ${}^wJ^t_B(q_0)$ 
denote the translational Jacobians of the contact points $A$ and $B$. 

In canonical inequality form, the contribution of pair $(a,b)$ is expressed as
\begin{equation*}
    \underbrace{d_{\min} - d_{AB}(q_0)}_{g_{ab}(q_0)}
    + \underbrace{{}^wn_{AB}^\top \left( {}^wJ^t_A(q_0) - 
    {}^wJ^t_B(q_0) \right)}_{G_{ab}(q_0)} 
    \, \Delta q \leq 0
\end{equation*}

Introducing such a constraint for every potential collision pair at each solver iteration 
would be computationally expensive.  
In practice, it is reasonable to assume that only bodies already in close proximity are likely 
to collide within a single control step.  
Therefore, an activation distance $d_{\text{active}} > d_{\min}$ is introduced:  
a pair $(a,b)$ is included in the \ac{QP} problem only if 
$d_{AB}(q_0) \leq d_{\text{active}}$. 

Aggregating all active pairs into the problem yields a constraint of the general form
\begin{equation}
    g_{\text{col}}(q_0) 
    + G_{\text{col}}(q_0) \, \Delta q \leq 0,
\end{equation}
with $g_{\text{col}}(q_0)$ and $G_{\text{col}}(q_0)$
obtained by stacking the contributions of all active pairs.

\section{Application to robot control}
\label{sec:applications}

This section illustrates the use of PlaCo kinematics solver through representative
examples. Each case highlights particular features of the solver and shows how 
different tasks and constraints can be seamlessly combined to achieve complex 
control objectives, while keeping the formulation straightforward to implement.

\begin{figure}[!ht]
    \centering
    \includegraphics[width=0.48\textwidth]{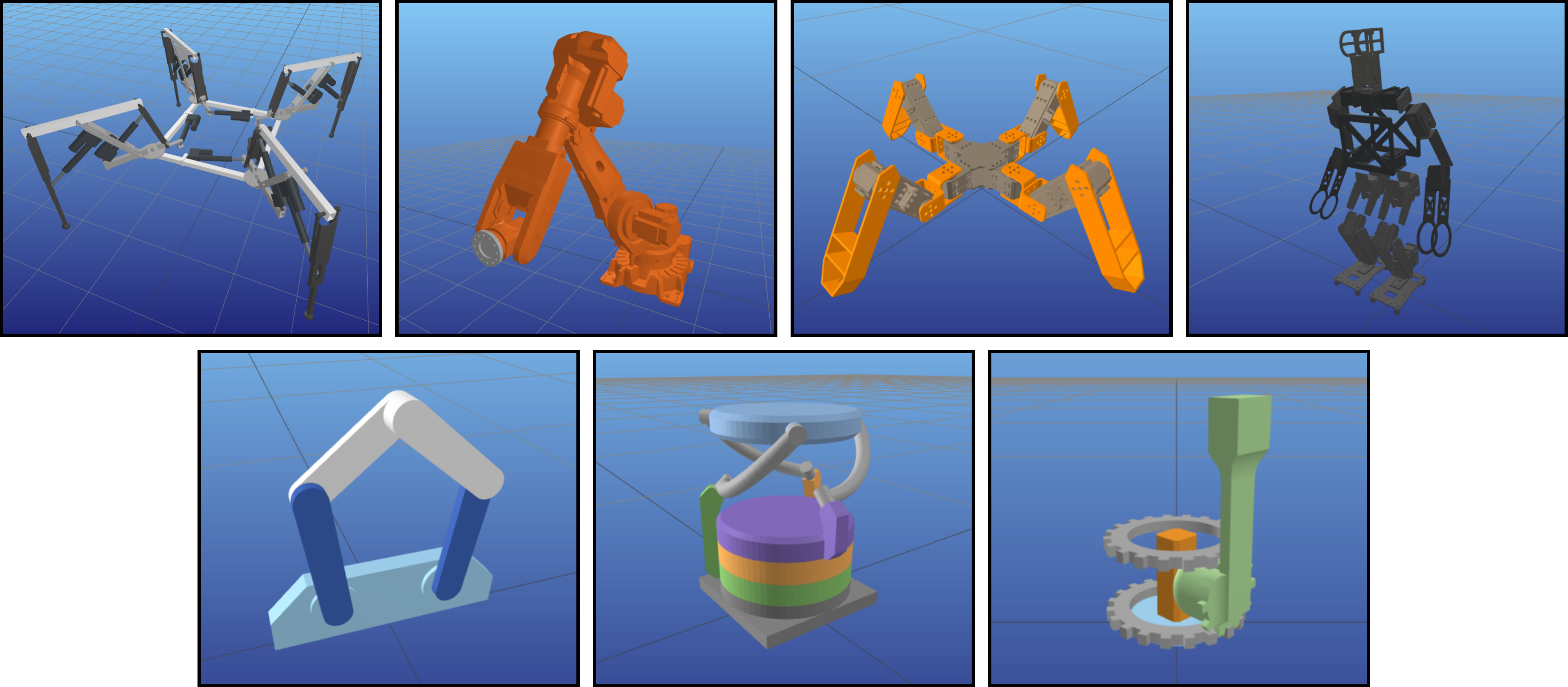}
    \vspace{-0.5em}
    \caption{Overview of the robot models available in the PlaCo example gallery\footref{fn:gallery}. 
    From top left to bottom right: a large quadruped
    with electric linear actuators and multiple kinematic loop-closures, 
    a 6-\ac{DoF} industrial manipulator mounted on a fixed base, 
    a small serial quadruped robot, 
    the Sigmaban humanoid platform, 
    a 2-\ac{DoF} planar arm containing a kinematic loop, 
    an orbita actuator~\cite{crampette2020orbita} with non-conventional joint arrangement, 
    and a differential joint mechanism.}
    \label{fig:placo-models}
    \vspace{-1em}
\end{figure}

A broader collection of examples, together with detailed source code, is available 
in the example gallery\footref{fn:gallery}.  
This gallery covers a wide range of robotic platforms, as illustrated in
\cref{fig:placo-models}. For each platform, both \ac{URDF} models and illustrative 
control scenarios are provided.

Three examples are detailed here. The first focuses on the control of a quadruped robot in 
static equilibrium, where multiple tasks and constraints are combined to maintain 
stability while enabling controlled motion. The second presents a kinematic loop-closure 
problem formulated with relative position tasks. Finally, the third demonstrates a differential joint, 
illustrating the use of the gear task to model complex mechanical transmissions.

\subsection{Quadruped balancing}
\label{sec:quadruped-example}

This first example illustrates a full-body inverse kinematics problem, in which 
a quadruped robot must maintain balance while simultaneously reaching for a 
target with one of its legs. The scenario demonstrates how multiple tasks and 
constraints, assigned with different priorities, can be combined within the solver 
to produce feasible and robust motions. 

The robot stands on three legs (\texttt{leg1}, \texttt{leg2}, and \texttt{leg3}), 
while the fourth leg (\texttt{leg4}) is tasked with reaching position targets.  
To ensure static equilibrium, the \ac{CoM} of the robot must remain 
within the support polygon defined by the stance legs. Additionally, the robot body 
is encouraged to remain in a predefined orientation, but is allowed to deviate if necessary to satisfy 
the other tasks. 

The first step is to fix the support legs at their respective positions. 
This is achieved by assigning hard-priority position tasks to 
\texttt{leg1}, \texttt{leg2}, and \texttt{leg3}: 

\begin{minted}[frame=single,fontsize=\scriptsize]{python}
# Adding hard-priority pos. tasks for the support legs
leg1 = solver.add_position_task("leg1", pos1)
leg2 = solver.add_position_task("leg2", pos2)
leg3 = solver.add_position_task("leg3", pos3)
leg1.configure("leg1", "hard", 1)
leg2.configure("leg2", "hard", 1)
leg3.configure("leg3", "hard", 1)
\end{minted}

\begin{figure}[!ht]
    \centering
    \includegraphics[width=0.48\textwidth]{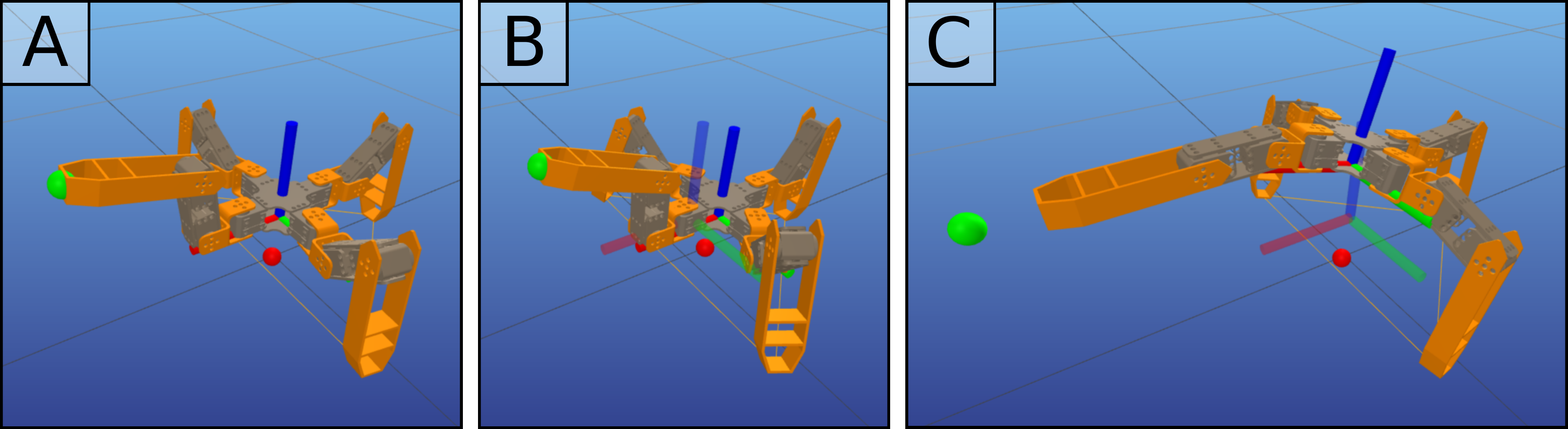}
    \vspace{-1em}
    \caption{Quadruped balancing on three legs while reaching with the fourth. 
    The green sphere marks the target for the moving leg, the red sphere the \ac{CoM} 
    projection, and the orange triangle the support polygon. 
    The current and desired body orientation are shown by the solid and transparent frames respectively.}
    \label{fig:placo-quad}
\end{figure}

To guarantee stability, a hard-priority \ac{CoM} polygonal constraint is introduced, 
forcing the horizontal projection of the \ac{CoM} to remain inside the support triangle 
formed by the stance legs:
\begin{minted}[frame=single,fontsize=\scriptsize]{python}
# Adding hard-priority CoM polygonal constraint
polygon = np.array([pos1, pos2, pos3])
com_const = solver.add_com_polygon_constraint(polygon)
com_const.configure("com_constraint", "hard", 1)
\end{minted}

A soft low-priority orientation task is then added for the robot body, 
encouraging it to remain upright but allowing deviations when necessary to 
satisfy higher-priority objectives:
\begin{minted}[frame=single,fontsize=\scriptsize]{python}
# Adding soft-low-priority orientation task for the body
T_world_body = tf.translation_matrix([0.0, 0.0, 0.05])
body_task = solver.add_frame_task("body", T_world_body)
body_task.configure("body", "soft", 1)
\end{minted}

Finally, a soft high-priority task is assigned to the free leg (\texttt{leg4}), 
driving it toward target positions $(x, y, z)$. The high weight ensures that
the reaching task is prioritized over the body orientation:
\begin{minted}[frame=single,fontsize=\scriptsize]{python}
# Adding soft-high-priority pos. task for leg4
target = np.array([x, y, z])
leg4 = solver.add_position_task("leg4", target)
leg4.configure("leg4", "soft", 1e3)
\end{minted}

\cref{fig:placo-quad} illustrates several typical outcomes of this scenario.  
In picture \textbf{A}, all tasks and constraints are simultaneously satisfied, and 
the quadruped maintains balance while reaching the target.  
In picture \textbf{B}, the target remains reachable but only at the expense of 
a deviation from the nominal body orientation, illustrating the trade-off between 
two soft tasks with different weights.  
Finally, picture \textbf{C} shows a situation where the target is unreachable: 
the solver prioritizes balance by keeping the \ac{CoM} inside the support polygon, while 
satisfying the reaching task as much as possible. \\~\vspace{-0.5em}

\vspace{-1em}

\subsection{Closed-loop Kinematics}
\label{sec:closed-loop}

Kinematic loops arise in various robotic systems, such as parallel manipulators 
or legged robots with closed-chain limbs. A conventional approach to 
controlling such systems consists in deriving an 
analytical relation between the actuated joints and the end-effector position 
and orientation. While this can be effective for a given architecture, the method 
lacks generality, since the relation depend on the specific geometry of the system. 
Moreover, this approach does not naturally handle joint limits or velocity bounds, 
which are critical in robotic applications. 

In PlaCo, closed-loop mechanisms are handled without requiring 
custom analytical derivations. Instead, the robot is modeled as an open 
kinematic chain in its \ac{URDF}, and loop-closure is enforced 
by introducing hard-priority relative position and/or orientation tasks.  
This approach preserves compatibility with the standard \ac{URDF} format 
while providing a flexible and generic solution to handle kinematic loops. 

\begin{figure}[b]
    \centering
    \includegraphics[width=0.48\textwidth]{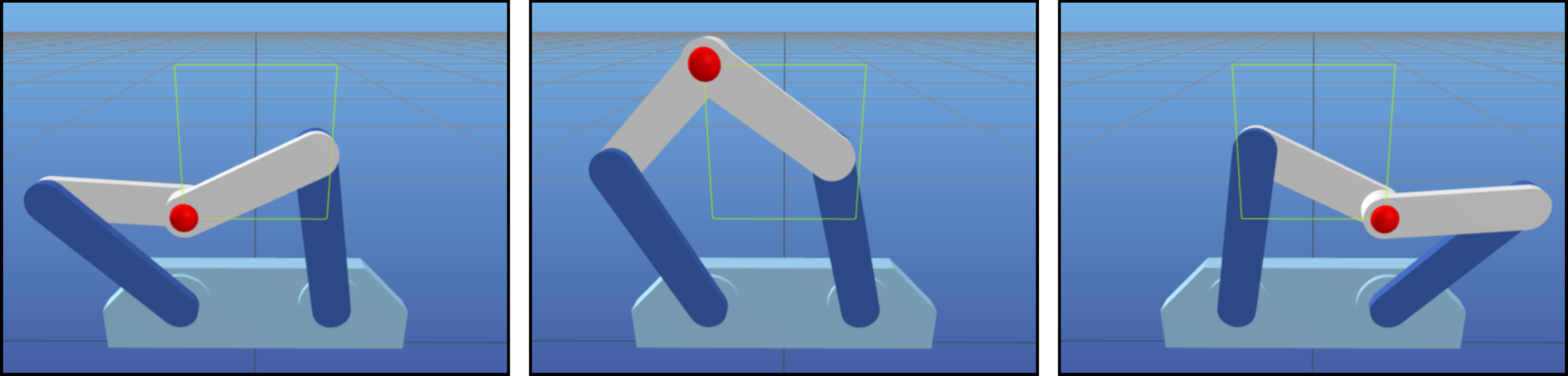}
    \vspace{-1em}
    \caption{Planar 2-\ac{DoF} robot with a kinematic loop, following a square trajectory with its end-effector (red).}
    \label{fig:placo-planar}
\end{figure}

To illustrate this method, consider the planar robot depicted in 
\cref{fig:placo-planar}. The robot consists of four segments, 
actuated only at its base by two motors, while the intermediate joints are passive. 
The goal is to control the position 
of the end-effector, shown in red, so that it follows a square trajectory. 

In the loaded \ac{URDF} model, the robot is represented as an open chain, 
with the two white terminal segments left unconnected.  
The kinematic loop is then closed by enforcing a hard-priority relative 
position task between points \texttt{c1} and \texttt{c2} located on these 
segments, constraining them to coincide: 
\begin{minted}[frame=single,fontsize=\scriptsize]{python}
# Adding loop closing task (hard priority)
closing_task = solver.add_relative_position_task(
    "c1", "c2", np.zeros(3)
)
closing_task.configure("closing", "hard", 1)
closing_task.mask.set_axises("xy")
\end{minted}

Note that only the $(x,y)$ components of the relative position task are 
constrained. Due to the planar nature of the mechanism, any error along 
the $z$ direction cannot be corrected through joint motions and is 
thus neglected by the solver. 

Once the loop is enforced, the end-effector can be controlled using a 
soft-priority position task, which drives it along a square trajectory in 
the $(x,z)$ plane: 
\begin{minted}[frame=single,fontsize=\scriptsize]{python}
# Adding a position task for the effector (soft priority)
target = np.array([x, 0.0, z])
end_task = solver.add_position_task("end", target)
\end{minted}

\subsection{Differential Joint}
\label{sec:differential}

Differential joints allow a single effective \ac{DoF} to be driven by the 
combined motion of two or more actuated joints. Such mechanisms are common in robotics, 
where they enable distributing loads across multiple actuators and 
provide mechanical advantages such as more compact designs. 

As for kinematic loops, a common approach to controlling differential mechanisms 
consists in deriving an analytical relation between the actuated input joints and 
the resulting motion of the differential outputs. However, this
method is confronted with the same limitations regarding generality and
the handling of joint limits and velocity bounds. 

\begin{figure}[!ht]
    \centering
    \includegraphics[width=0.48\textwidth]{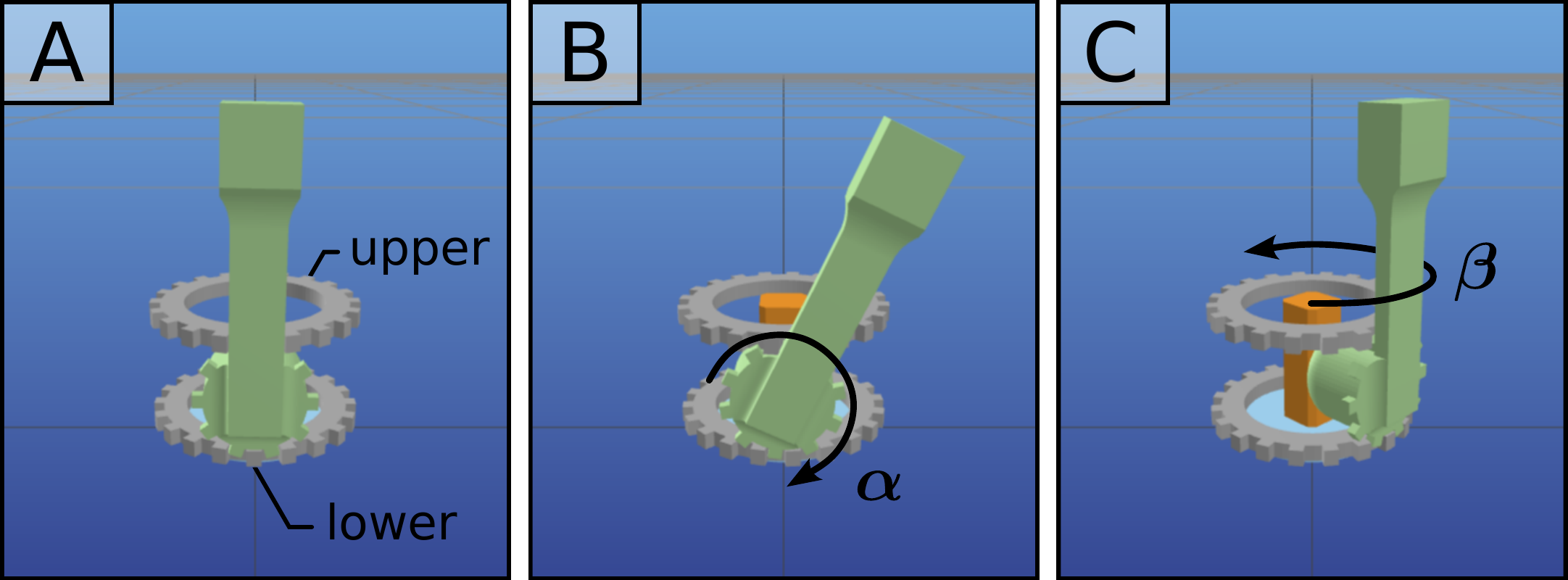}
    \vspace{-1em}
    \caption{Differential joint mechanism. The two active \acp{DoF}, upper and lower, are represented 
    by the gears in the model, as presented in picture \textbf{A}. 
    Picture \textbf{B} illustrates a pure $\alpha$ motion, obtained by rotating 
    the actuated \acp{DoF} in opposite directions.  
    Picture \textbf{C} illustrates a pure $\beta$ motion, obtained by rotating 
    the actuated \acp{DoF} in the same direction.}
    \label{fig:differential}
\end{figure} 

PlaCo offers a more versatile solution by modeling such mechanisms through a 
gear task, which directly encodes the linear coupling between active and passive 
joints. This task provides a unified formalism to represent a wide variety of 
mechanical transmissions, independently of their underlying architecture, while 
seamlessly integrating with other tasks and constraints in the \ac{QP} problem. 

Consider the differential mechanism shown in \cref{fig:differential}, composed 
of two active joints \texttt{upper} and \texttt{lower}, and two passive joints 
$\alpha$ and $\beta$. Using a gear task, one can enforce the couplings:
\begin{equation*}
    \alpha = \texttt{upper} - \texttt{lower}, 
    \qquad 
    \beta = \tfrac{1}{2} (\texttt{upper} + \texttt{lower}),
\end{equation*}
which correspond respectively to the relative and average motion of the two actuated joints.  
The implementation in PlaCo is straightforward:
\begin{minted}[frame=single,fontsize=\scriptsize]{python}
# Adding gears coupling constraints (hard priority)
gear_task = solver.add_gear_task()
gear_task.configure("gear", "hard", 1)
gear_task.add_gear("alpha", "upper", 1)
gear_task.add_gear("alpha", "lower", -1)
gear_task.add_gear("beta", "upper", 0.5)
gear_task.add_gear("beta", "lower", 0.5)
\end{minted}

Once the coupling constraints are established, the mechanism can be controlled 
either at the level of the actuated joints or directly through the passive joints.  
Prescribing values $(x, y)$ for the active joints corresponds to solving the direct 
kinematics of the differential, while specifying targets $(a, b)$ for the passive joints 
$\alpha$ and $\beta$ amounts to solving its inverse kinematics:
\begin{minted}[frame=single,fontsize=\scriptsize]{python}
# Adding a joint task to control (soft priority)
joints_task = solver.add_joints_task()

# Control via active joints
joints_task.set_joints({"lower": x, "upper": y})

# Control via passive joints
joints_task.set_joints({"alpha": a, "beta": b})
\end{minted}

In both cases, the solver enforces the gear constraints automatically and 
respects any specified joint position or velocity limits on the actuated 
joints \texttt{upper} and \texttt{lower}, ensuring both feasibility and safety.

% This example illustrates how the gear task allows differential mechanisms to be 
% handled in a modular and general way, without the need for architecture-specific 
% analytical derivations, and while benefiting from the full set of features of the \ac{QP} solver. 

\section{Conclusion}
\label{sec:placo-conclusion}

This article presents PlaCo, a software framework that streamlines 
the formulation and solution of \ac{QP}-based planning and control problems 
for robotics.

\Cref{sec:placo-architecture} first presents the overall architecture of PlaCo. 
Two of the main components are then described in detail. The problem formulation
layer, covered in \cref{sec:problem-formulation}, provides a high-level interface 
that abstracts away the low-level matrix assembly required for \ac{QP} 
problem definition. This interface supports weighted objectives, distinguishes 
between hard and soft constraints, 
and incorporates integrated decision variables that embed linear dynamics 
directly into the optimization. Building on this foundation, the
whole-body kinematics solver of PlaCo is detailed in \cref{sec:kinematics-solver}.
This module supports a broad set of task types -- such as joint regulation 
and absolute or relative frame control -- and complementary constraints, including 
polygonal feasibility regions and self-collision avoidance. 

The versatility of the solver is showcased through three representative
examples in \cref{sec:applications}. These examples illustrated how PlaCo 
can combine heterogeneous tasks and constraints with different priorities 
to produce feasible and robust behaviors, while keeping the modeling 
effort modest. 

PlaCo is open-source and available on GitHub\footnote{Repository: \href{https://github.com/Rhoban/placo}
{github.com/Rhoban/placo}}. The Python API is 
distributed via pip, allowing straightforward installation. 
Its design philosophy is to decouple model 
description from numerical implementation, thereby lowering the barrier for 
adopting \ac{QP}-based control across diverse robotic platforms. 
The development of comparable tools, such as the contemporary 
pink library~\cite{pink}, further suggests that PlaCo addresses a genuine 
need within the robotics community.

% \subsection{PlaCo adoption}

% PlaCo has seen growing adoption within the robotics community. 
% The repository currently counts around 200 GitHub stars, and several open-source 
% projects have integrated the library. Notably, it is used in Hugging Face–sponsored 
% initiatives such as the Open Duck Mini project, the Reachy Mini platform developed 
% by Pollen Robotics, and the LeRobot project~\cite{cadene2024lerobot}. \\

% According to PyPI statistics, the package was 
% downloaded more than 44{,}000 times between March and 
% August~2025, mirrors excluded. 
% This corresponds to approximately 250 downloads per day on average, 
% reflecting sustained interest from practitioners and researchers alike. \\

% PlaCo has also been cited in the scientific literature, for instance 
% in~\cite{xu2025dexumi}, which attests to its relevance within the 
% robotics research community.

\subsection{Perspectives}

% While PlaCo focuses on a lightweight and efficient core centered on \ac{QP} formulations, 
% its scope could be broadened through alternative backends. For example, CasADi~\cite{Andersson2019} 
% would provide access to additional \ac{QP} solvers and enable seamless extensions to nonlinear 
% programming, thereby covering problems that go beyond the expressive power of convex quadratic models. \\

An ongoing line of development is the \ac{WPG} module presented in \cref{fig:placo-architecture}, which implements a 
walking pattern generator based on the \ac{LIPM}. This module is already functionnal, 
but improving its documentation and usability remains a work in progress.

Another promising direction is the integration of a whole-body dynamics solver. 
This module, already present in the repository, is designed to extend the abstraction 
principles of the kinematics solver to dynamics-based control. Although the details of 
this work fall outside the scope of the present manuscript, it points toward a more 
comprehensive framework that unifies kinematics and dynamics within the same high-level interface.

% Finally, given the community uptake and to facilitate proper scholarly citation, 
% preparing a dedicated article is a natural next step.

\appendix
\renewcommand{\thesection}{\Alph{section}}
\setcounter{section}{0}

\noindent{\hypertarget{sec:integrated-example}
{\textit{A. Problem formulation usage}}}\par
\vspace{0.5em}
\label{sec:usage-example}

The concepts introduced in \cref{sec:problem-formulation} 
-- weighted objectives, hard and soft constraints, 
and integrated decision variables -- are encapsulated within the
\texttt{Problem} class of PlaCo. This class provides a high-level interface for
constructing optimization problems.

A problem is defined by attaching a set of decision variables, on which
constraints and objectives can be specified. Internally, these elements are
automatically translated into the \ac{QP} form of
equation~\eqref{eq:qp}, and solved efficiently using a dedicated quadratic
programming backend. In PlaCo, this is implemented in C++ through the
EiQuadProg solver~\cite{eiquadprog_soft}, ensuring real-time performance. The class
exposes both a C++ API for online control and Python bindings for rapid
prototyping.

Decision variables can also be associated with an \texttt{Integrator},
which embeds them into a linear dynamical model as described in
\cref{sec:integrator}. By default, the integrator implements a
chain of integrators of arbitrary order, as in the example presented in 
\hyperlink{sec:integrated-example}{Appendix B},
where the \ac{CoM} jerk served as decision variable. More generally, any linear model
can be specified by providing the continuous-time state matrix $D$.

A simple illustrative example in python is provided below to demonstrate the use of
the \texttt{Problem} and \texttt{Integrator} classes. It shows how decision
variables can be defined, integrated, and constrained in a concise and
declarative manner.

\begin{minted}[frame=single,fontsize=\scriptsize]{python}
# Definition of a problem with jerk as decision variable
problem = placo.Problem()
dddx = problem.add_variable(10)
integ = placo.Integrator(dddx, np.zeros(3), 3, 0.1)

# Intermediate waypoint constraints on position
problem.add_constraint(integ.expr(3, 0) <= -0.5)
problem.add_constraint(integ.expr(7, 0) >=  1.5)

# Terminal constraints
problem.add_constraint(integ.expr(10,0) == 1.0)
problem.add_constraint(integ.expr(10,1) == 0.0)
problem.add_constraint(integ.expr(10,2) == 0.0)

# Solving the underlying QP problem
problem.solve()
\end{minted}

In this example, a third-order integrator chain is created with a sampling
period of 0.1\,s and zero initial conditions for position, velocity, and
acceleration. The decision variable is a vector of 10 piecewise-constant
jerks, corresponding to a one-second horizon. Constraints are imposed on
the position at intermediate steps 3 and 7, as well as on position,
velocity, and acceleration at the final step 10. By default, problem constraints 
are treated as hard, but they can be specified as soft by calling the 
\texttt{configure()} method and providing a weight.

\begin{figure}[!ht]
    \centering
    \includegraphics[width=0.48\textwidth]{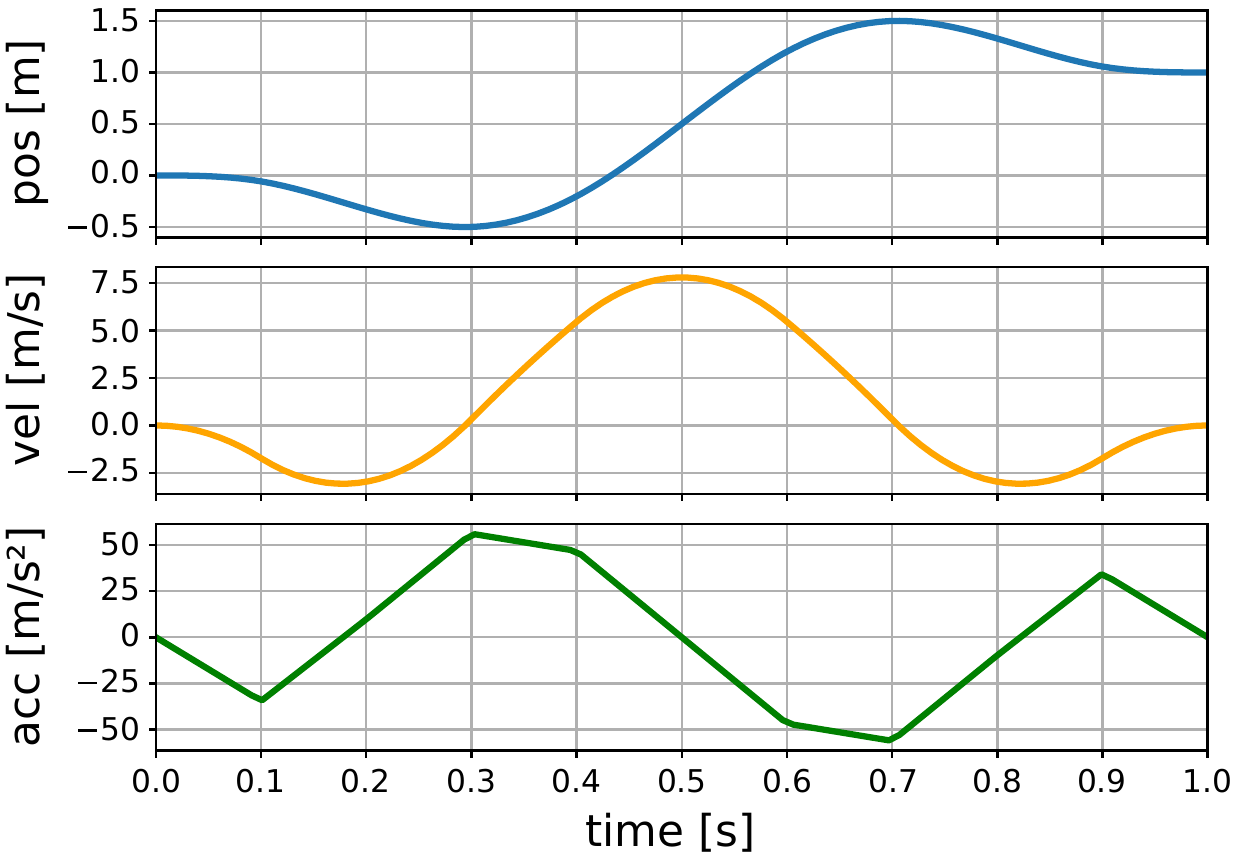}
    \caption{Trajectories obtained by solving the problem defined in the code snippet.}
    \label{fig:placo-problem-result}
\end{figure}

The corresponding trajectories, obtained by solving the problem, are shown in
\cref{fig:placo-problem-result}. This example highlights the design
philosophy of PlaCo: low-level \ac{QP} formulation is fully abstracted, while the
user specifies only the meaningful constraints of the control
problem.

\noindent{\hypertarget{sec:integrated-example}
{\textit{B. Integrated decision variables -- CoM example}}}\par
\vspace{0.5em}

% \paragraph{Example -- \ac{CoM} dynamics with \ac{CoM} jerk as decision variable}

Consider the continuous-time dynamics of the \acf{CoM} of a robot:
\begin{equation}
    \begin{bmatrix}
        \dot{c} \\
        \ddot{c} \\
        \dddot{c}
    \end{bmatrix}
    = 
    \underbrace{
    \begin{bmatrix}
        0 & 1 & 0 \\
        0 & 0 & 1 \\
        0 & 0 & 0
    \end{bmatrix}
    }_{D}
    \begin{bmatrix}
        c \\
        \dot{c} \\
        \ddot{c}
    \end{bmatrix}
    +
    \underbrace{
    \begin{bmatrix}
        0 \\
        0 \\
        1
    \end{bmatrix}
    }_{E}
    \dddot{c}
    \label{eq:com_dyn}
\end{equation}

By choosing the jerk $\dddot{c}$ as decision variable, a \ac{QP}
problem can be formulated in which the \ac{CoM} position, velocity, and acceleration
can be constrained. The derivation of the discrete-time dynamics follows the
procedure outlined in \cref{sec:integrator}:
\begin{equation}
    \hspace{-0.8em}
    \exp\!\left(
    \begin{bmatrix}
    D & E \\
    0 & 0
    \end{bmatrix}
    \Delta t
    \right)
    =
    \begin{bmatrix}
    \overbrace{
    \begin{bmatrix}
    1 & \Delta t & \tfrac{1}{2}\Delta t^2 \\
    0 & 1 & \Delta t \\
    0 & 0 & 1
    \end{bmatrix}
    }^{D_{d}}
    &
    \overbrace{
    \begin{bmatrix}
    \tfrac{1}{6}\Delta t^3 \\
    \tfrac{1}{2}\Delta t^2 \\
    \Delta t
    \end{bmatrix}
    }^{E_{d}} \\
    0 & 1
    \end{bmatrix}
\end{equation}

This yields the discrete-time dynamics:
\begin{equation}
    \begin{bmatrix}
        c_{k+1} \\
        \dot{c}_{k+1} \\
        \ddot{c}_{k+1}
    \end{bmatrix}
    = 
    \begin{bmatrix}
    1 & \Delta t & \tfrac{1}{2}\Delta t^2 \\
    0 & 1 & \Delta t \\
    0 & 0 & 1
    \end{bmatrix}
    \begin{bmatrix}
        c_k \\
        \dot{c}_k \\
        \ddot{c}_k
    \end{bmatrix}
    +
    \begin{bmatrix}
    \tfrac{1}{6}\Delta t^3 \\
    \tfrac{1}{2}\Delta t^2 \\
    \Delta t
    \end{bmatrix}
    \dddot{c}_k
    \label{eq:discrete_com_dyn}
\end{equation}
\vspace{0em}

Knowing the matrices $D, E, D_{d}, E_{d}$, 
one can constrain the \ac{CoM} position, velocity, and acceleration at any time $t$ of the horizon
as a linear function of the jerk decision variables.

\bibliographystyle{ieeetr}
\bibliography{biblio}

\end{document}